\documentclass[runningheads,a4paper]{llncs}

\usepackage{amssymb}
\usepackage{graphicx}
\usepackage{booktabs}
\usepackage{graphicx}
\usepackage{amsmath,amsfonts,amssymb}
\usepackage[tight]{subfigure}
\usepackage{flushend}
\usepackage[active]{srcltx}

\begin{document}
\title{Assessment of SAR Image Filtering using Adaptive Stack Filters}
\author{Mar\'ia E. Buemi\inst{1} \and Marta Mejail \inst{1}\and Julio Jacobo \inst{1} \and A.\ C.\ Frery \inst{2} \and H.\ S.\ Ramos \inst{2}}
 \institute{Departamento de Computaci\'on\\
  	Facultad de Ciencias Exactas y Naturales\\
  	Universidad de Buenos Aires
 \and
 LCCV \& CPMAT\\
  	Instituto de Computa\c c\~ao\\
  	Universidade Federal de Alagoas
 }
\maketitle

\thispagestyle{empty}

\begin{abstract}
Stack filters are a special case of non-linear filters.
They have a good performance for filtering images with different types of noise while preserving edges and details.
A stack filter decomposes an input image into several binary images according to a set of thresholds.
Each binary image is then filtered by a Boolean function, which characterizes the filter.
Adaptive stack filters can be designed to be optimal; they are computed from a pair of images consisting of an ideal noiseless image and its noisy version.
In this work we study the performance of adaptive stack filters when they are applied to Synthetic Aperture Radar (SAR) images.
This is done by evaluating the quality of the filtered images through the use of suitable image quality indexes and by measuring the classification accuracy of the resulting images.
\end{abstract}


\section{Introduction}\label{introduc}

SAR images are generated by a coherent illumination system and are affected by the coherent interference of the signal from the terrain~\cite{OliverQuegan98}.
This interference causes fluctuations of the detected intensity which varies from pixel to pixel, an effect called speckle noise, that also appears in ultrasound-B, laser and sonar imagery.
Speckle noise, unlike noise in optical images, is neither Gaussian nor additive; it follows other distributions and is multiplicative.
Classical techniques, therefore, lead to suboptimal results when applied to this kind of imagery.
The physics of image formation leads to the following model: the observed data can be described by the random field $Z$, defined as the product of two independent random fields: $X$, the backscatter, and $Y$, the speckle noise.
The backscatter is a physical magnitude that depends on the geometry and water content of the surface being imaged, as well as on the angle of incidence, frequency and polarization of the electromagnetic radiation emitted by the radar.
It is the main source of information sought in SAR data.
Different statistical distributions have been proposed in the literature for describing speckled data.
In this work, since we are dealing with intensity format, we use the Gamma distribution, denoted by $\Gamma$, for the speckle, and the reciprocal of Gamma distribution, denoted by $\Gamma^{-1}$, for the backscatter.
These assumptions, and the independence between the fields, result in the intensity $\mathcal{G}^{0}$  law for the return~\cite{frery96}.
This family of distributions is indexed by three parameters: roughness $\alpha$, scale $\gamma$, and the number of looks $n$, and it has been validated as an universal model for several types of targets.
Speckle has a major impact on the accuracy of classification procedures, since it introduces a low signal-to-noise ratio.
The effectiveness of techniques for combating speckle can be measured, among other quantities, through the accuracy of simple classification methods.
The most widespread statistical classification technique is the Gaussian maximum likelihood classifator.
Stack filters are a special case of non-linear filters.
They have a good performance for filtering images with different types of noise while preserving edges and details.
Some authors have studied these filters, and many methods have been developed for their construction and applicaton as in~\cite{Prasad2005}.
These filters decompose the input image, by thresholds, in binary slices.
Each binary image is then filtered using a Boolean function evaluated on a sliding window.
The resulting image is obtained summing up all the filtered binary images.
The main drawback in using stack filters is the need to compute optimal Boolean functions.
Direct computation on the set of all Boolean functions is unfeasible, so most techniques rely on the use of a pair of images: the ideal and corrupted one.
The functions are sought to provide the best estimator of the former using the latter as input.
The stack filter design method used in this work is based on an algorithm proposed by Yoo \textit{et al.}~\cite{YooKelvinHuangCoyleAdams}.

We study the application of this type of filter to SAR images, assessing its performance by evaluating the quality of the filtered images through the use of image quality indexes like the universal image quality index and the correlation measure index and by measuring the classification accuracy of the resulting images using maximum likelihood Gaussian classification.

The structure of this paper is as follows: In Section~\ref{sec:imagSAR} we summarise the $\mathcal{G}^{0}$ model for speckled data. Section~\ref{define_stack} gives an introduction to stack filters, and describes the filter design method used in this work. In Section~\ref{sec:Resultados} we discuss the results of filtering through image quality assesment and classification performance.
Finally, in Section~\ref{sec:Conclusiones} we present the conclusions.

\section{The Multiplicative Model}\label{sec:imagSAR}

Following~\cite{Petty:LAAR:ProtocoloLee}, we will only present the univariate intensity case.
Other formats (amplitude and complex) are treated in detail in~\cite{frery96}.

The intensity $\mathcal G^0$ distribution that describes speckled return is characterized by the following density:
\begin{equation*}
f(z)=\frac{L^{L}\Gamma(L-\alpha)}{\gamma^{\alpha}\Gamma(L)\Gamma(-\alpha)}\frac{z^{L-1}}{(\gamma+Lz)^{L-\alpha}},
\end{equation*}
where $-\alpha,\gamma,z>0$, $L\geq1$, denoted $\mathcal G^{0}(\alpha,\gamma,L)$.

The $\alpha$ parameter corresponds to image roughness (or heterogenity).
It adopts negative values, varying from $-\infty$  to $0$.
If $\alpha$ is near $0$, then the image data are extremely heterogeneous (for example: urban areas), and if $\alpha$ is far from the origin then the data correspond to a homogeneous region (for example: pasture areas). The values for forests lay in-between.

Many filters have been proposed in the literature for combating speckle noise, among them the ones by Lee and by Frost.
These filters will be applied to speckled data, along with the filter proposed in this work. For quality performance the comparision will be done between the stack filter and the Lee filter. Classification performance will be assessed by classifying data filtered with the Lee, Frost and stack filters using a  Gaussian maximum likelihood approach.

\section{Stack Filters}\label{define_stack}

This section is dedicated to a brief synthesis of stack filter definitions and design.
For more details on this subject, see~\cite{YooKelvinHuangCoyleAdams,AstolaKuosmanen,LinKim}.

Consider images of the form $X\colon S\rightarrow \{0,\dots,M\}$, with $S$ the support and $\{0,\dots,M\}$ the set of admissible values.
The threshold is the set of operators $T^{m}\colon \{0,\dots,M\}\rightarrow\{0,1\}$ given by 
\begin{equation*}
T^{m}(x)=\left\{ 
\begin{array}{ccc}
1 & \text{if} & x\geq m, \\ 
0 & \text{if} & x < m.
\end{array}
\right.   \label{def_umbral}
\end{equation*}
We will use the notation
$X^{m}=T^{m}(x)$.  
According to this definition, the value of a non-negative integer number $x\in \{0,\dots,M\}$ can be reconstructed making the summation of its thresholded values between $0$ and $M$. 
Let $X=(x_{0},\ldots,x_{n-1})$ and $Y=(y_{0},\ldots,y_{n-1})$ be binary vectors of length $n$, define an order relation given by
$X\leq Y$ if and only if holds that $x_{i}\leq y_{i}$ for every $i$.
This relation is reflexive, anti-symmetric and transitive, generating therefore a partial ordering on the set of binary vectors of fixed length.
A boolean function $f\colon\{ 0,1\}^{n}\rightarrow\{0,1\}$, where $n$ is the length of the input vectors, has the stacking property if and only if 
\begin{equation*}
\forall X,Y\in\{0,1\}^{n},\ X\leq Y\Rightarrow f(X)
\leq f( Y) .  \label{prop_stacking}
\end{equation*}

We say that $f$ is  a positive boolean function if and only if it can be written by means of an expression that contains only non-complemented input variables.
That is,
$f( x_{1},x_{2},\ldots ,x_{n})=\bigvee_{i=1}^{K} \bigwedge_{j\in P_{i}}x_{j},
\label{form_func_bool_positiva}$
where $n$ is the number of arguments of the function, $K$ is the number of terms of the expression and $P_{i}$ is a subset of the interval $\{1, \ldots, N\}$. `$\bigvee$' and `$\bigwedge$' are the AND and OR Boolean operators.
It is possible to proof that this type of functions has the stacking property.

A stack filter is defined by the function $S_{f}\colon \{0, \ldots, M\}^{n}\rightarrow \{0, \ldots, M \} $, corresponding to the Positive Boolean function $f(x_{1}, x_{2}, \ldots, x_{n}) $ expressed in the given form by~(\ref{form_func_bool_positiva}).
The function $S_{f}$ can be expressed by means of
$S_{f}( X) =\sum_{m=1}^{M}f( T^{m}( X)) .
\label{form_reconstruccion_equivalencia}$

In this work we applied the stack filter generated with the fast algorithm described in \cite{YooKelvinHuangCoyleAdams}.

Stack filters are built by a training process that generates a positive boolean function that preserves the stacking property.
Originally, this training is performed providing two complete images on $S$, one degraded and one noiseless.
The algorithm seeks the operator that best estimates the later using the former as input, and as a means of measuring error.

The implementation developed for this work supports the application of the stack filter many times. Our approach consists of using a set of regions of interest, much smaller than the whole data set, and relying on the analysis the user makes of these information.
Graphical and quantitative analyses are presented. 
The user is prompted with the mean value of each region as the default desired value, but he/she can choose other from a menu (including the median, the lower and upper quartiles and a free specification).
This freedom of choice is particularly useful when dealing with non-Gaussian degradation as is the case of, for instance, impulsive noise.

\section{Results}\label{sec:Resultados}

In this section, we present the results of building stack filters by training. These filters are applied to both simulated and real data. %
The stack filters obtained are compared to SAR image filters. This comparision is done by assessing smoothing and edge preservation through image quality indexes and by evaluating the influence of filtering on classification performance.
\subsection{Image quality assesment}\label{sec:CalidadDeImagen}

The indexes used to evaluate the quality of the filtered images are the universal image quality index \cite{WangBovik}  and the correlation measure $\beta$.
The universal image quality index $Q$ is given byequation (\ref{form_WangBovikIndex})
\begin{equation}		
Q = \frac{\sigma_{XY}}{\sigma_{X} \sigma_{Y}} \frac{2\overline{X}\overline{Y}}{\overline{X}^2+\overline{Y}^2}
\frac{2\sigma_{X}\sigma_{Y}}{\sigma_{X}^2+\sigma_{Y}^2},
\label{form_WangBovikIndex}
\end{equation}
where $\sigma_{X}^2 = (N-1)^{-1}\Sigma_{i=1}^N (X_i - \overline{X})^2$, $\sigma_{Y}^2 = (N-1)^{-1}\Sigma_{i=1}^N (Y_i - \overline{Y})^2$, $\overline{X} = N^{-1}\Sigma_{i=1}^N X_i$ and $\overline{Y} = N^{-1}\Sigma_{i=1}^N Y_i$.
The dynamic range of index $Q$ is $[-1,1]$, being $1$ the best value. To evaluate the index of the whole image, local indexes $Q_i$ are calculated for each pixel using a suitable square window, and then these results are averaged to yield the total image quality $Q$.
The correlation measure is given by
\begin{equation}
\beta = \frac{\sigma_{\nabla^{2}X\nabla^{2}Y}}{\sigma^2_{\nabla^{2}X} \sigma^2_{\nabla^{2}Y}}, 
\label{form_Beta}
\end{equation}
where $\nabla^{2}X$ and $\nabla^{2}Y$ are the Laplacians of images $X$ and $Y$, respectively. 

\begin{table}[hbt]
\caption{Statistics from image quality indexes}\label{tab:imagequality}
    \centering
    \begin{tabular}{ c||rr||rr||rr||rr}\toprule
	    & \multicolumn{4}{c||}{$\beta$ index} & \multicolumn{4}{c}{$Q$ index} \\ \midrule
	       &       \multicolumn{2}{c||}{Stack filter}  & \multicolumn{2}{c||}{Lee filter} &       \multicolumn{2}{c||}{Stack filter}  & \multicolumn{2}{c}{Lee filter}\\ \midrule
	    contrast & \multicolumn{1}{c}{$\overline{\beta}$} & \multicolumn{1}{c||}{$s_\beta$} &     \multicolumn{1}{c}{$\overline{\beta}$} & \multicolumn{1}{c||}{$s_\beta$} & \multicolumn{1}{c}{$\overline{Q}$} & \multicolumn{1}{c||}{$s_Q$} & \multicolumn{1}{c}{$\overline{Q}$} & \multicolumn{1}{c}{$s_Q$}   \\ \hline
	    10:1	& 0.1245 & 0.0156  &    0.0833  & 0.0086   & 0.0159 & 0.0005 &     0.0156 & 0.0004 \\
	    10:2	& 0.0964 & 0.0151  &    0.0663  & 0.0079   & 0.0154 & 0.0005 &     0.0148 & 0.0004  \\
	    10:4	& 0.0267 & 0.0119  &    0.0421 &  0.0064   & 0.0124  & 0.0008 &     0.0120 & 0.0006\\
	    10:8	& $-$0.0008 & 0.0099  &   0.0124   & 0.0064  & 0.0041 & 0.0013 &     0.0021 & 0.0006  \\ \bottomrule

        \end{tabular}
\end{table}

In Table \ref{tab:imagequality} the correlation measure $\beta$ and the quality index $Q$ are shown. The comparison is made between Lee filtered and   stack filtered SAR images. To this end, a Monte Carlo experiment was performed, generating 1000 independent replications of synthetic 1-look SAR images for each of four contrast ratios. The generated images consist of two regions separated by a vertical straight border. Each sample corresponds to a different contrast ratio, wich ranges from $10$:$1$ to $10$:$8$. This was done in order to study the effect of the contrast ratio in the quality indexes considered.

It can be seen that, according to the results obtained for the $\beta$ index, the stack filter exhibits a better performance at high contrast ratios, namely $10$:$1$ and $10$:$2$, while the Lee filter shows the opposite behavior.
The results for the $Q$ index show slightly better results for the stack filter all over the range of contrast ratios. It is remarkable the small variance of these estimations, compared to the mean values obtained.

\begin{figure}[htb]
\centering
\subfigure[Values of $\beta$, Lee filter\label{fig:betaLee}]{\includegraphics[width=0.48\linewidth]{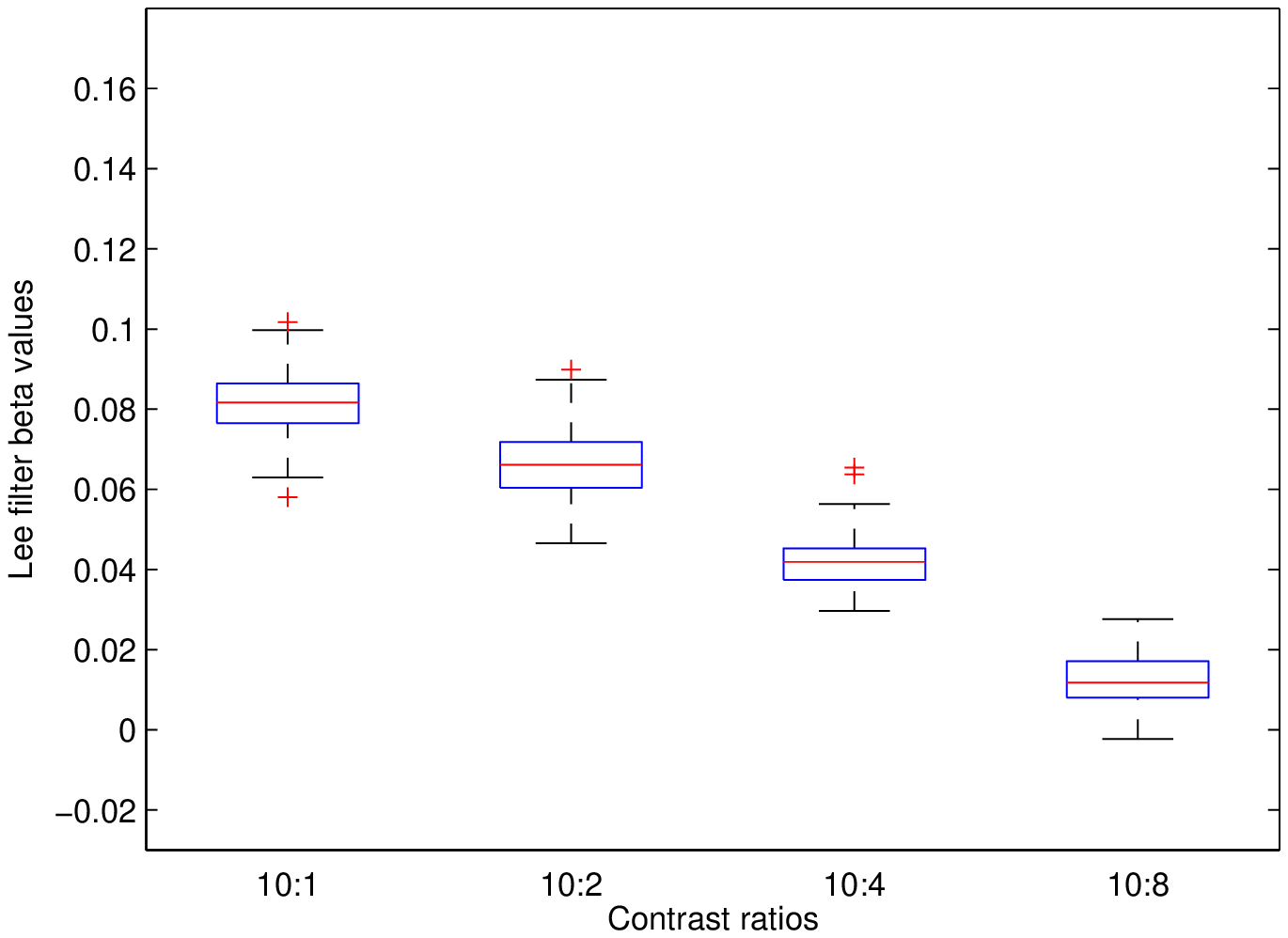}} 
\subfigure[Values of $\beta$, Stack filter\label{fig:betaStack}]{\includegraphics[width=0.48\linewidth]{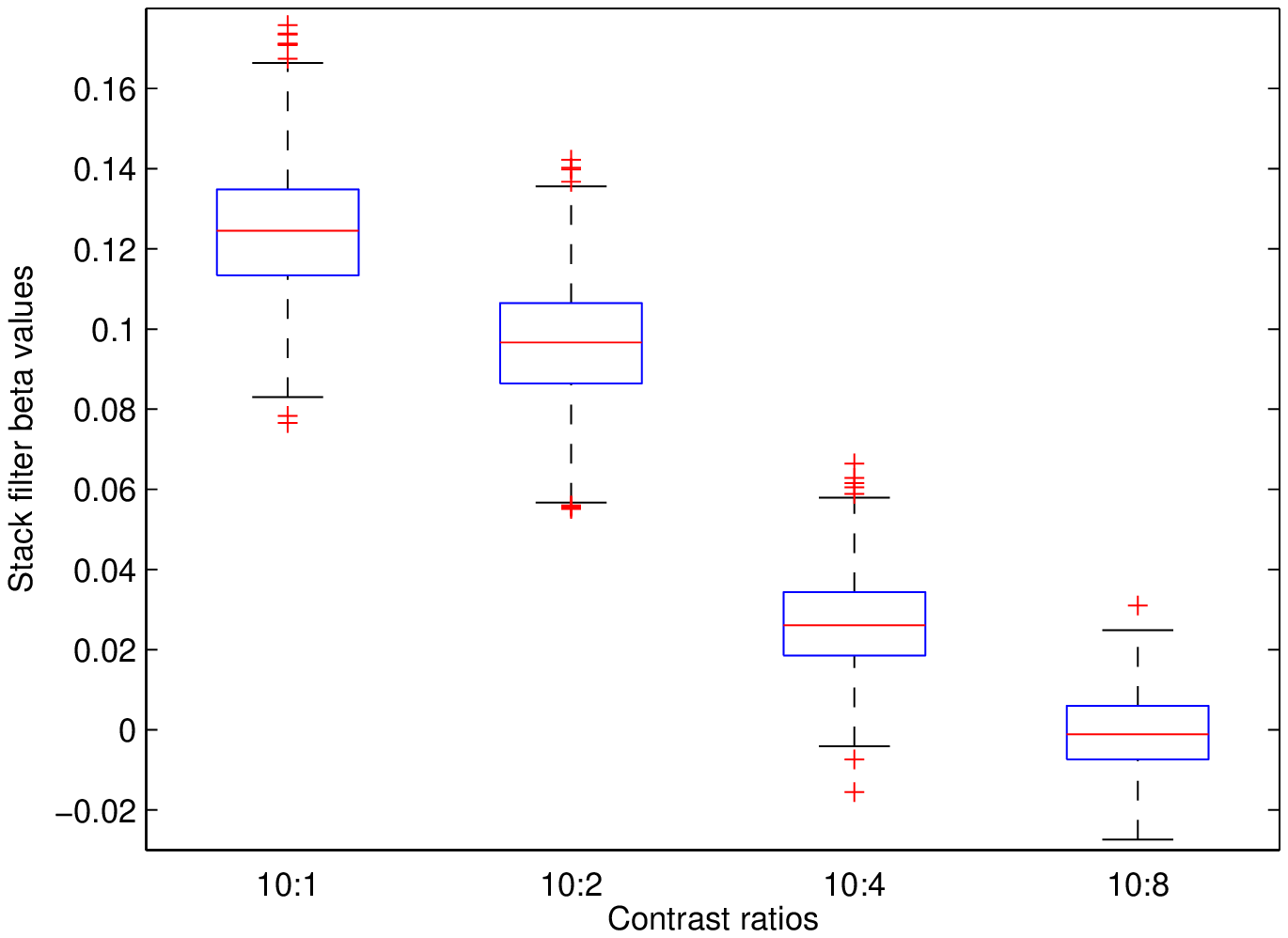}}
\subfigure[Values of $Q$, Lee filter\label{fig:QLee}]{\includegraphics[width=0.48\linewidth]{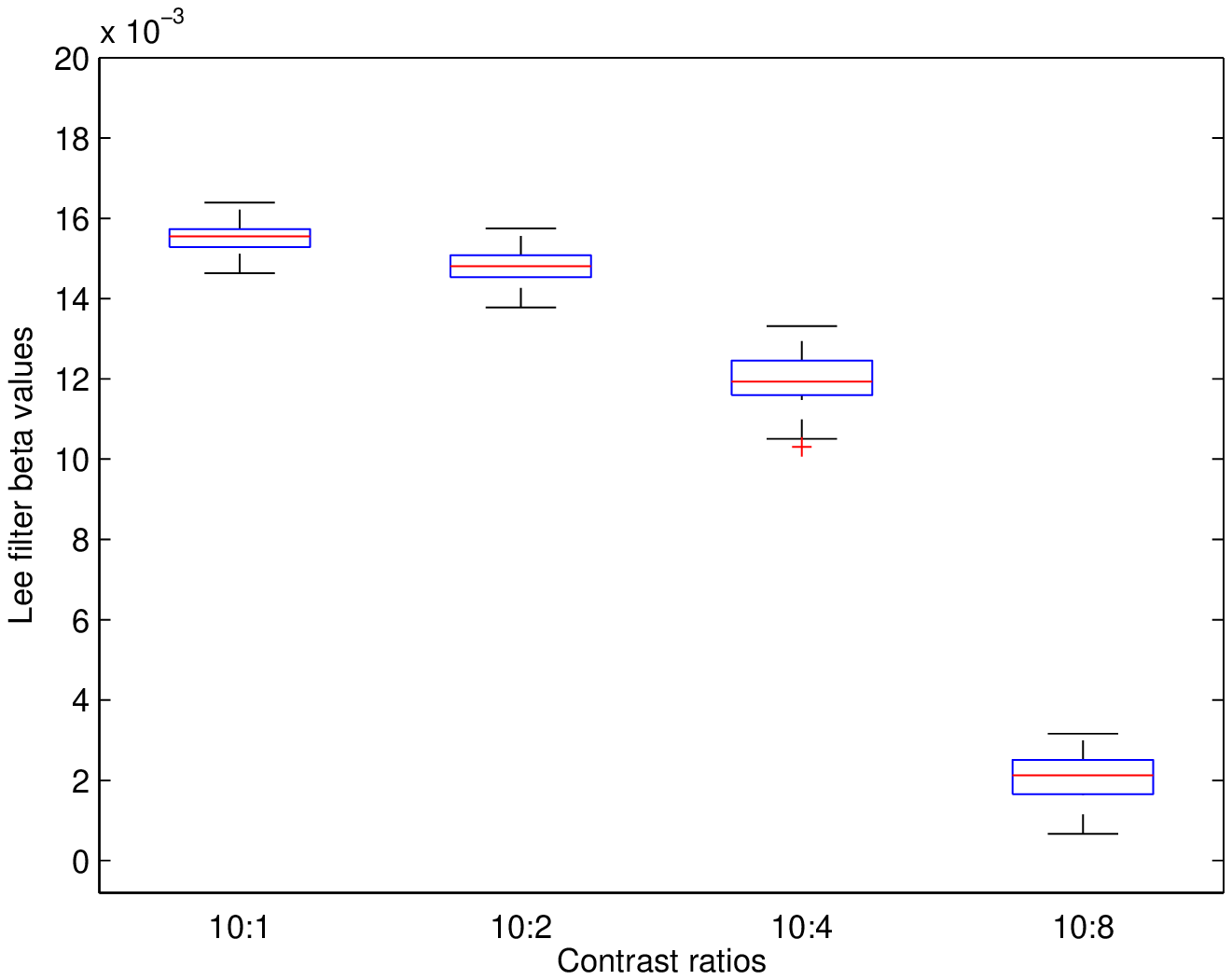}}
\subfigure[Values of $Q$, Stack filter\label{fig:QStack}]{\includegraphics[width=0.48\linewidth]{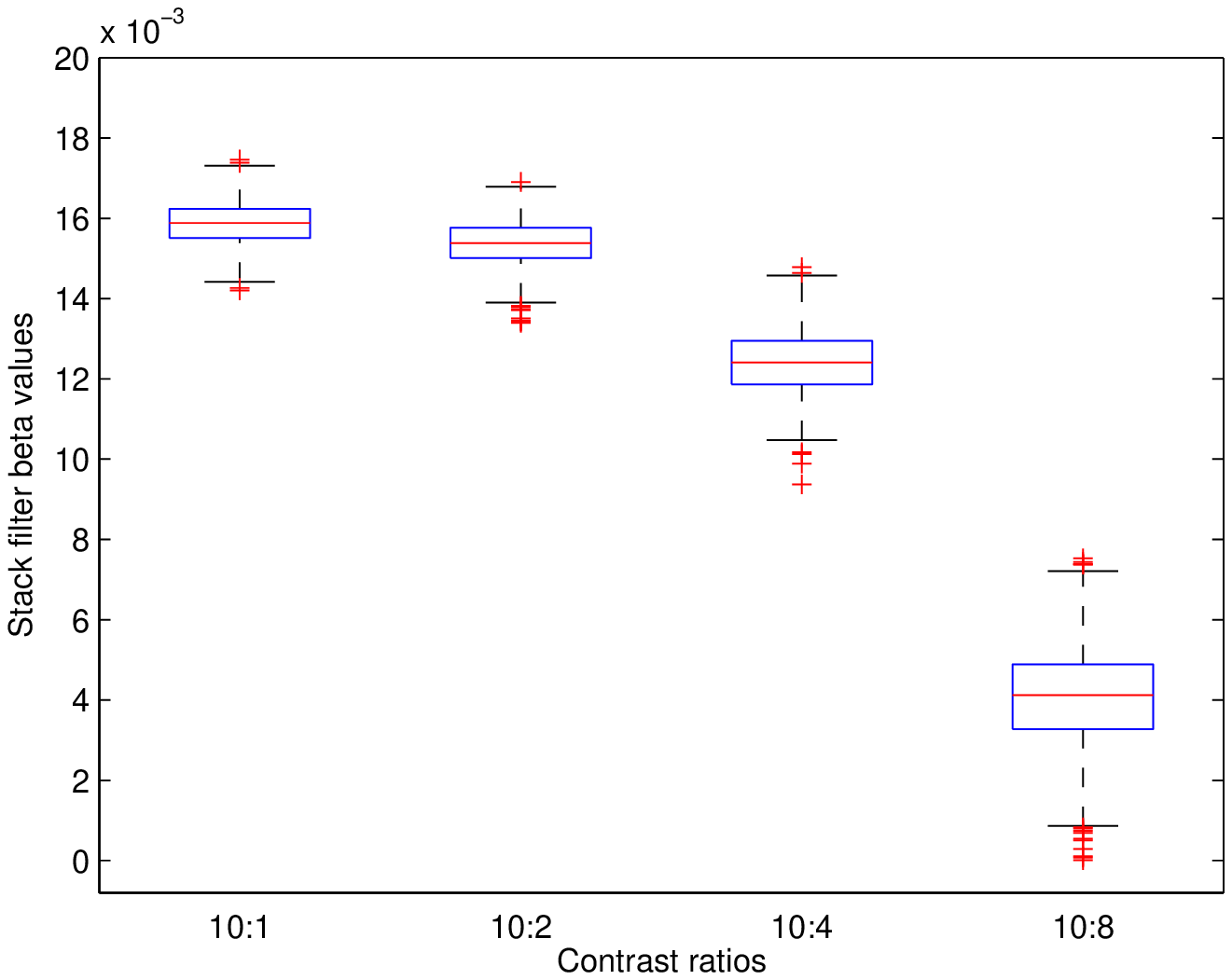}}
\caption{Boxplots of the quality indexes}
\label{fig:boxplots}
\end{figure}

Fig.~\ref{fig:boxplots} shows the boxplots of the observations summarized in Table~\ref{tab:imagequality}. From the plots for the $\beta$ index, it can be seen that, the Lee filter has a lower degree of variability with contrast and that both are almost symmetric.  The plots of the $Q$ index show a better performance for the stack filter for all the contrast ratios considered.

\subsection{Classification performance}\label{sec:SegmentationPerformance}

The equality of the classification results are obtained by calculating the confusion matrix, after Gaussian Maximum Likelihood Classification (GMLC).

\begin{figure*}[htb]
  \begin{center}
\subfigure[Simulated image and GMLC\label{fig:simulated2reg}]{\includegraphics[width=.14 \linewidth]{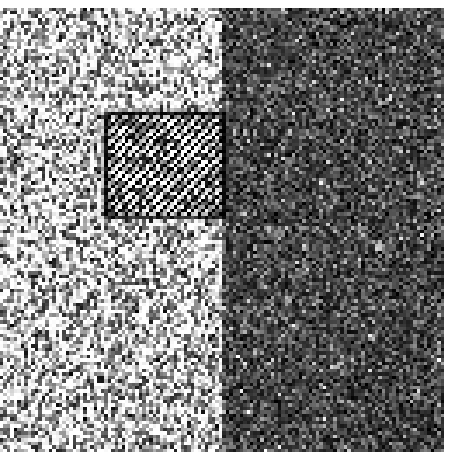}
\includegraphics[width=.14 \linewidth]{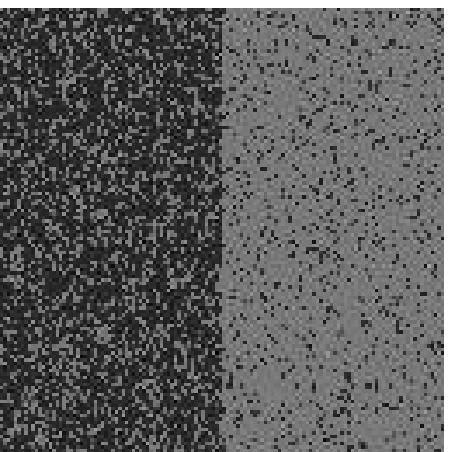}} 
\quad
\subfigure[One iteration and GMLC\label{fig:1itersimul}]{ \includegraphics[width=.14 \linewidth]{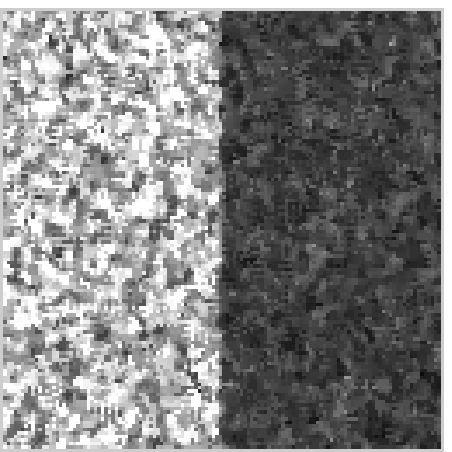}
\includegraphics[width=.14 \linewidth]{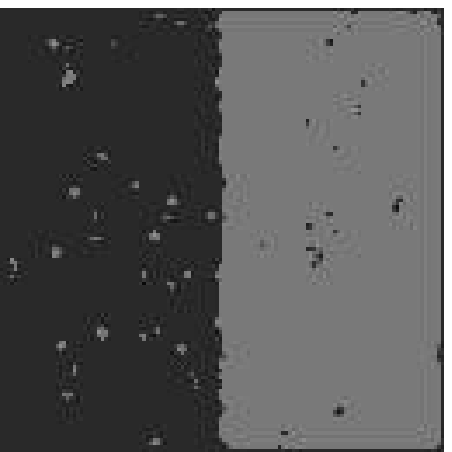}}
\quad
\subfigure[$95$ iterations and GMLC\label{fig:95itersimul}]{\includegraphics[width=.14 \linewidth]{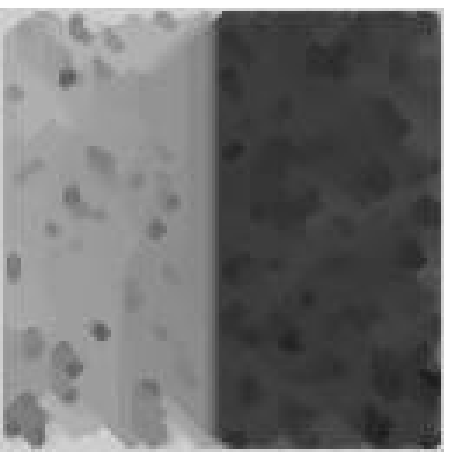}
\includegraphics[width=.14 \linewidth]{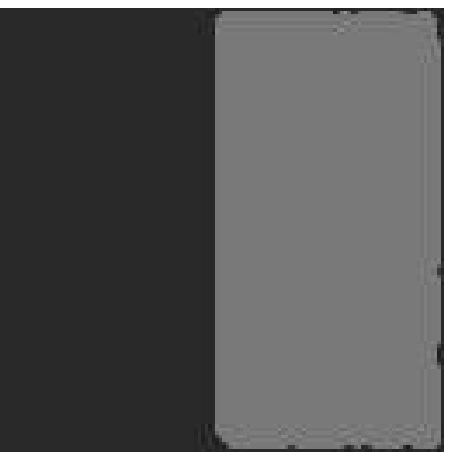}} 
 \end{center}
 \caption{Training by region of interest: simulated data}
\label{fig:training_simulated}
\end{figure*}

Fig.~\ref{fig:simulated2reg}, left, presents an image $128\times128$ pixels, simulated with two regions: samples from the $\mathcal G^0(-1.5,\gamma^*_{-1.5,1},1)$ and from the $\mathcal G^0(-10,\gamma^*_{-10,1},1)$ laws form the left and right halves, respectively, where $\gamma^*_{\alpha,n}$ denotes the scale parameter that, for a given roghness $\alpha$ and number of looks $n$ yields an unitary mean law.
In this manner, Fig.~\ref{fig:simulated2reg} presents data that are hard to classify: extremely heterogeneous and homogeneous areas with the same mean, with the lowest possible signal-to-noise ratio ($n=1$).
The mean value of the dashed area was used as the ``ideal'' image.
Fig.~\ref{fig:1itersimul} and~\ref{fig:95itersimul}, left, show the result of applying the resulting filter once and $95$ times, respectively.
The right side of Fig.~\ref{fig:simulated2reg}, Fig.~\ref{fig:1itersimul} and Fig.~\ref{fig:95itersimul} present the GMLC of each image.
Not only the pointwise improvement is notorious, but the edge presevation is also noteworthy, specially in Fig.~\ref{fig:95itersimul}, right, where the straight border has been completely retrieved.

\begin{figure*}[htb]
  \begin{center}
 \subfigure[Image, samples and GMLC\label{fig:realimage}]{ \includegraphics[width=.15 \linewidth]{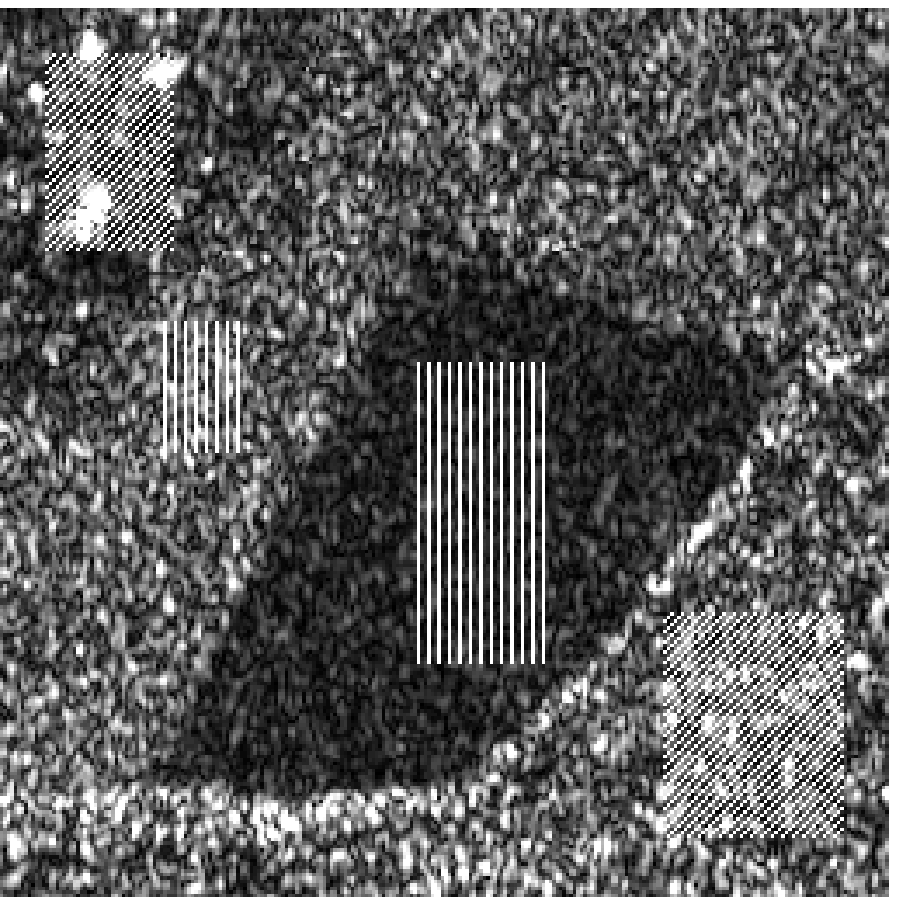}
\includegraphics[width=.15 \linewidth]{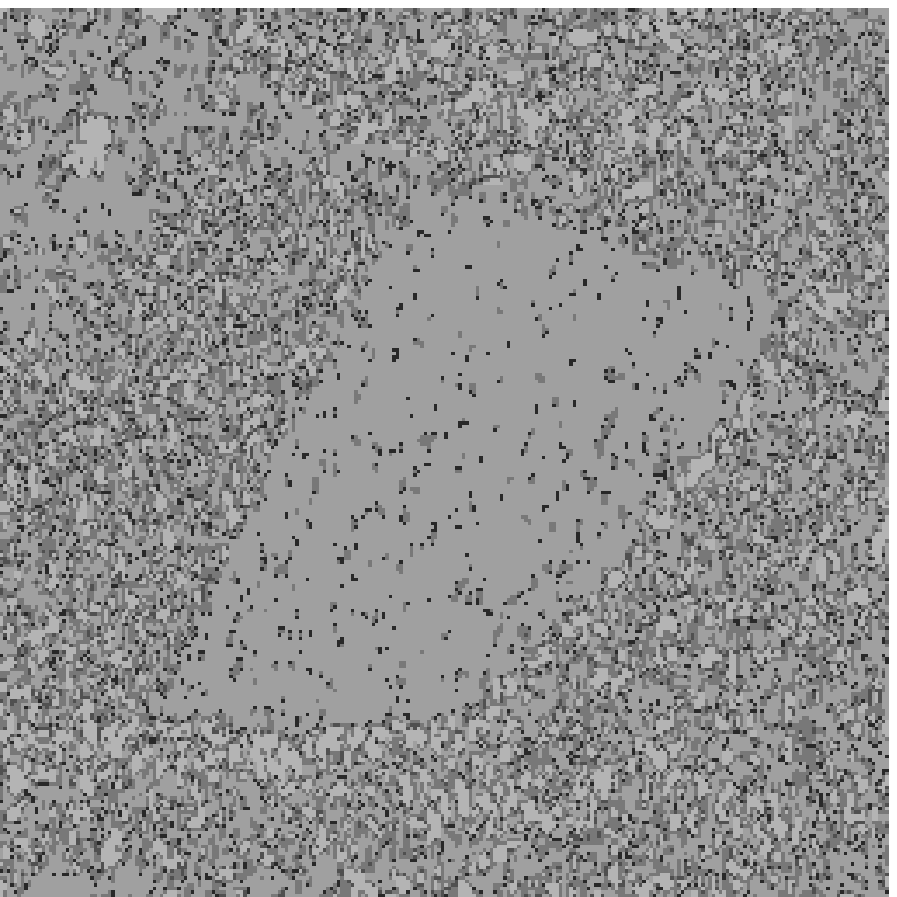}}
 \subfigure[Frost and GMLC\label{fig:filfrost}]{\includegraphics[width=.15 \linewidth]{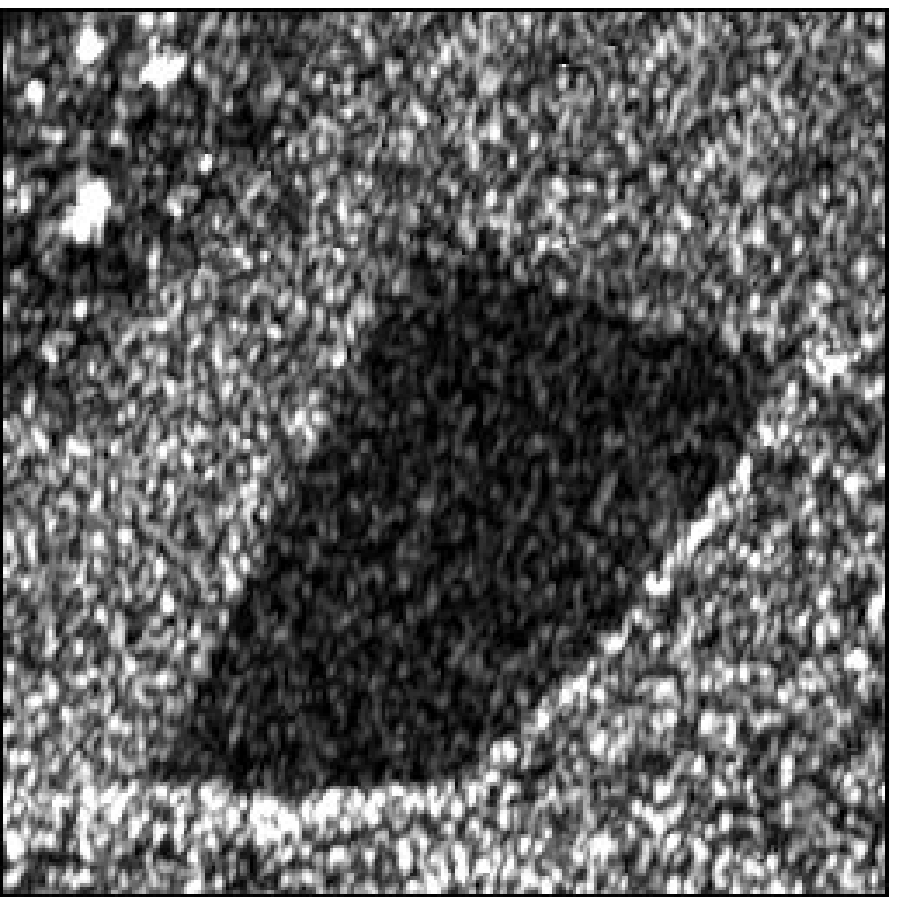}
\includegraphics[width=.15\linewidth]{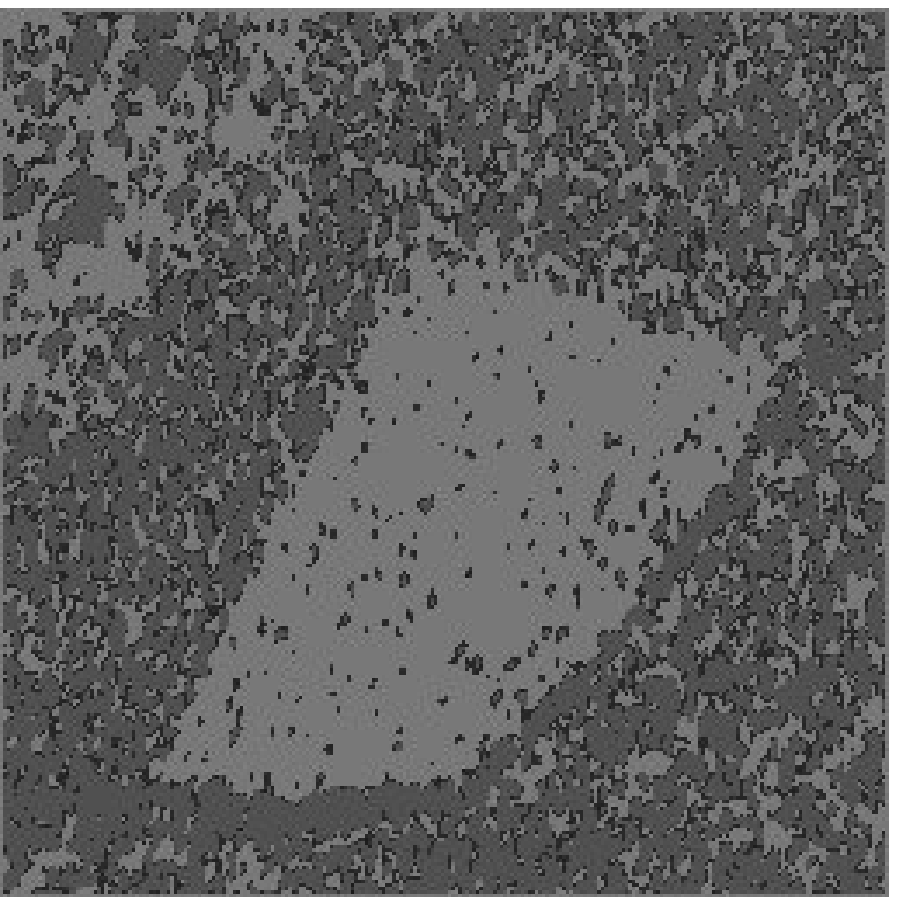}}
 \subfigure[Lee and GMLC\label{fig:fillee}]{\includegraphics[width=.15 \linewidth]{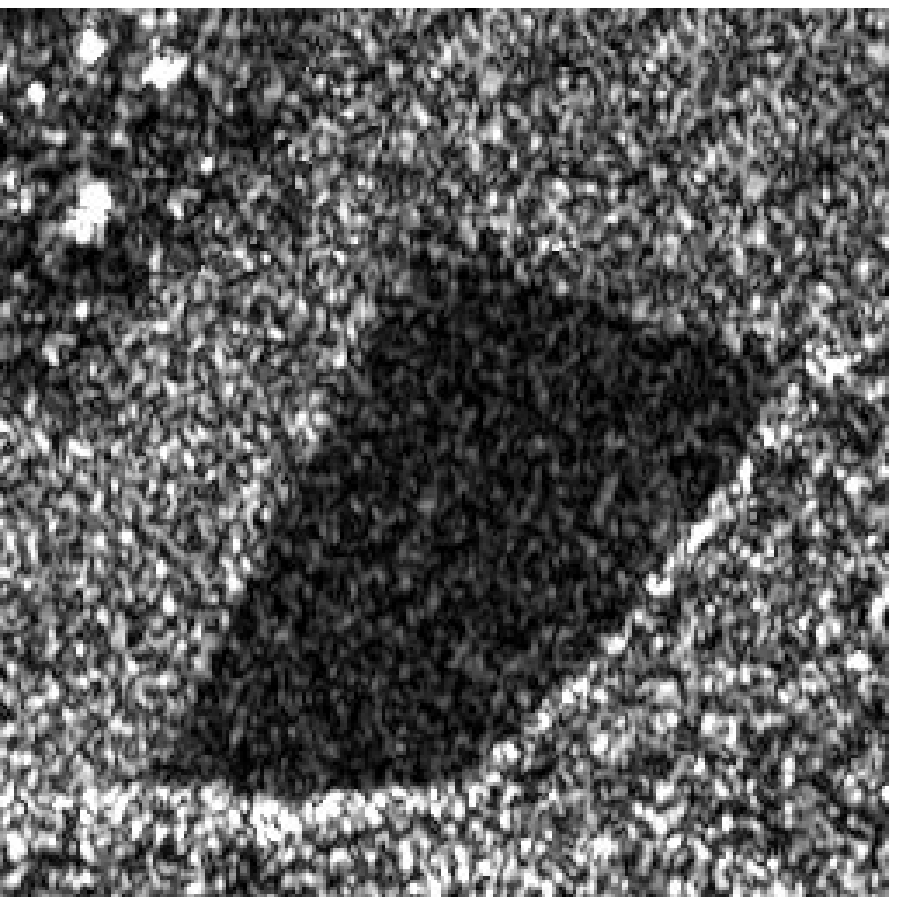}
 \includegraphics[width=.15 \linewidth]{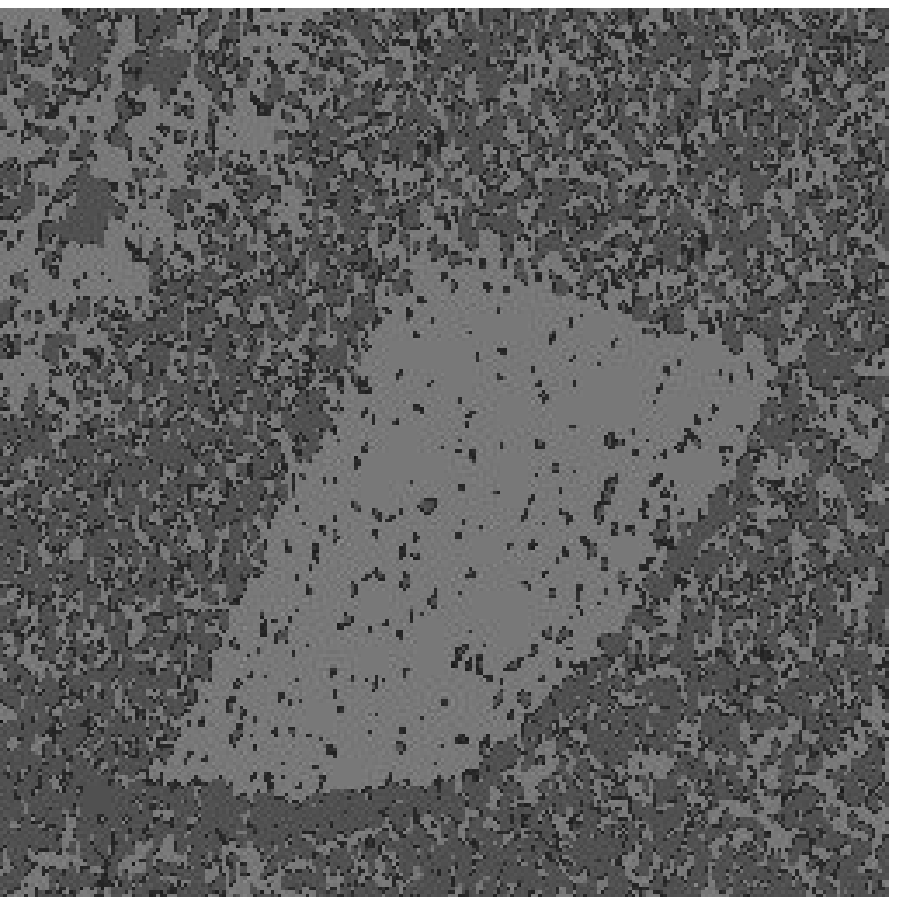}}
\subfigure[Stack Filter $20$ and GMLC \label{fig:stackfilter95}]{\includegraphics[width=0.15\linewidth]{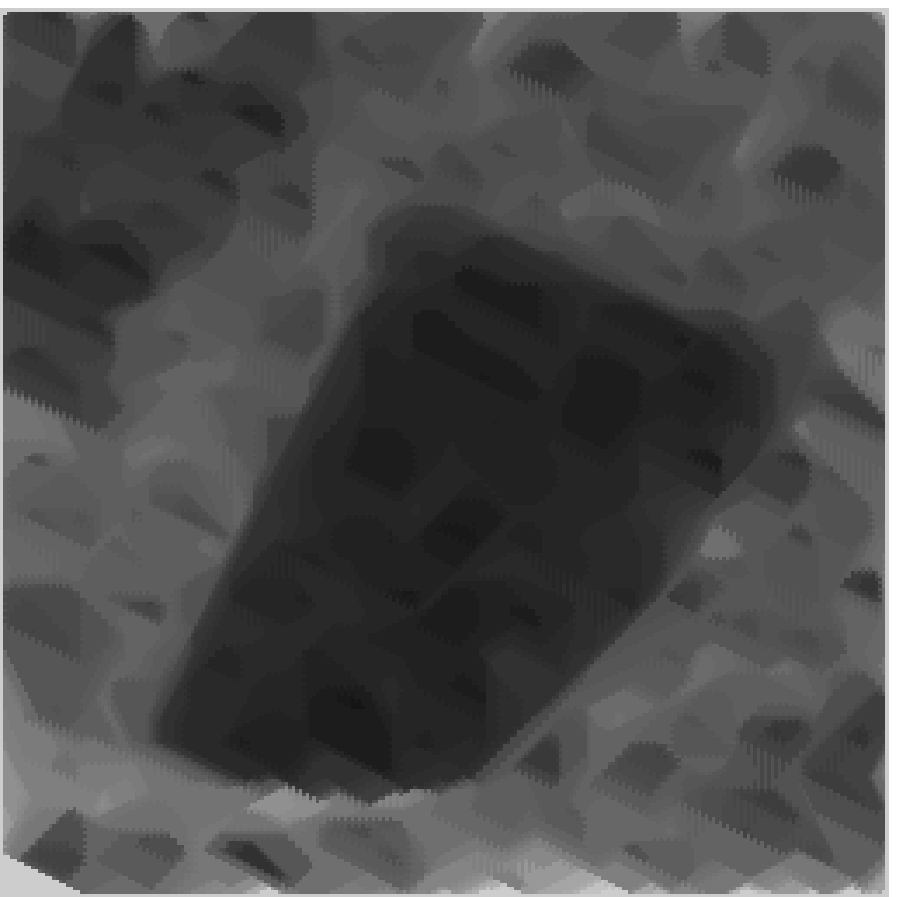}
\includegraphics[width=0.15\linewidth]{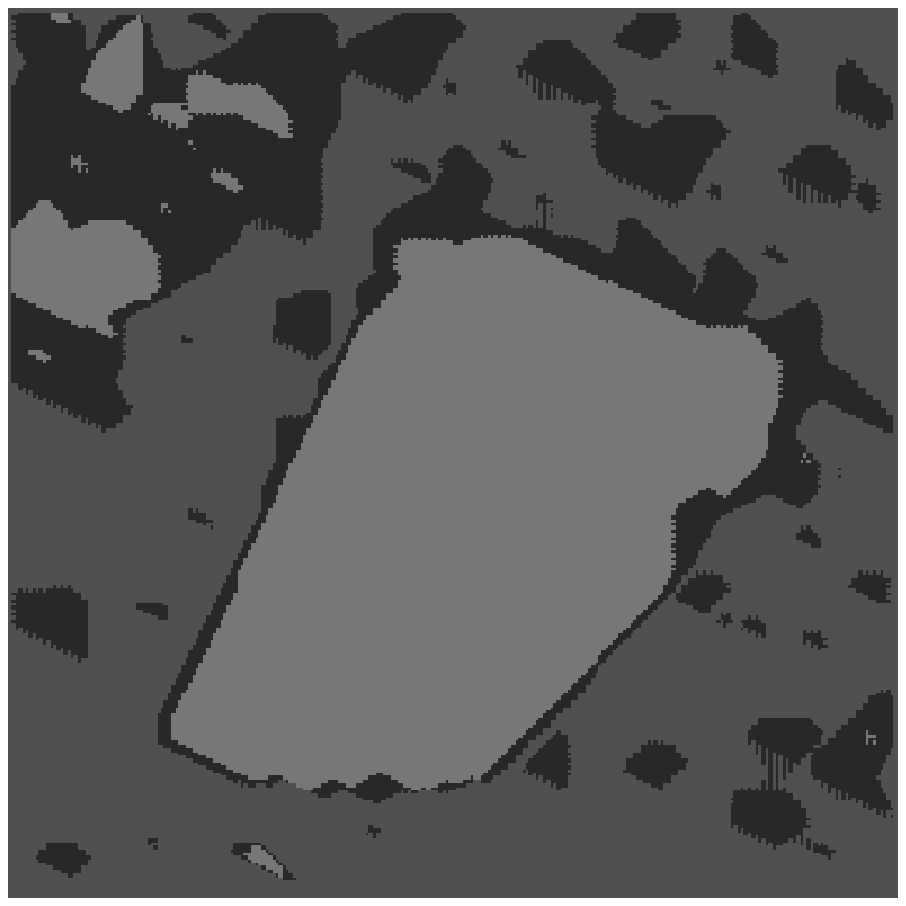}}
\end{center}
 \caption{Training by region of interest: real image}\label{fig:visual}
\end{figure*}

Fig.~\ref{fig:visual} compares the performance of the proposed stack filter with respect to two widely used SAR filters: Lee and Kuan.
Fig.~\ref{fig:realimage} presents the original data, and the regions of interest used for estimating the Boolean function.
In this case, again, the mean on each region was used as the `ideal' image.
Fig.~\ref{fig:filfrost}, Fig.\ref{fig:fillee} and Fig.~\ref{fig:stackfilter95} present the result of applying the Frost, Lee and Stack filters (one and $22$ iterations) to the original SAR data.
The right side of previous figures present the corresponding GMLC.
The stack filter produces better results than classical despeckling techniques.

Table~\ref{tab:confusion223} presents the main results from the confusion matrices of all the GMLC, including the results presented in~\cite{Buemi:Sibgrapi:2007} which used the classical stack filter estimation with whole images.
It shows the percentage of pixels that was labeled by the user as from region $R_i$ that was correctly classified as belonging to region $R_i$, for $1\leq i\leq 3$.
``None'' denotes the results on the original, unfiltered, data, ``Sample Stack $k$'' denotes our proposal of building stack filters with samples, applied $k$ times, ``Stack $k$'' the classical construction applied $k$ times, and ``Frost'' and ``Lee'' the classical speckle reduction filters.

It is clear the superior performance of stack filters (both classical and by training) over speckle filters, though the stack filter by training requires more than a single iteration to outperform the last ones.

Stack filters by training require about two orders of time less than classical stack filters to be built, and they produce comparable results.
Using regions of interest is, therefore, a competitive approach.

\begin{table}[htb]
\caption{Statistics from the confusion matrices}\label{tab:confusion223}
    \centering
    \begin{tabular}{ c rrr }\toprule
            Filter & $R_1$/$R_1$ & $R_2$/$R_2$ & $R_3$/$R_3$   \\ \midrule
                   None &13.40 &48.16 &88.90\\ \midrule
	            Sample Stack 1   & 9.38 &65.00 &     93.19\\
        	    Sample Stack 22  & \textbf{63.52} &\textbf{74.87} & \textbf{96.5}\\ \midrule
                    Stack 1  &14.35  &64.65 &90.86\\
                    Stack 40 &62.81  &89.09 &94.11\\
                    Stack 95 &\textbf{63.01} &\textbf{93.20} &\textbf{94.04}\\ \midrule
                    Frost &16.55 &55.54 &90.17\\
                    Lee   &16.38 &52.72 &89.21\\ \bottomrule
        \end{tabular}
\end{table}
\section{Conclusions} \label{sec:Conclusiones}

In this work, the effect of adaptive stack filtering on SAR images was assessed. Two viewpoints were considered: a classification performance viewpoint and a quality perception viewpoint. For the first approach, the Frost and Lee filters were compared with the iterated stack filter using a metric extracted from the confusion matrix. A real SAR image was used in this case.
For the second approach, a Monte Carlo experience was carried out in which 1-look synthetic SAR, i.e., the noisiest images, were generated. In this case, the Lee filter and a one pass stack filter were compared for various degrees of contrast. The $\beta$ and the $Q$ indexes were used as measures of perceptual quality.
The results of the $\beta$ index shows that the stack filter performs better in cases of high contrast.
The results of the $Q$ index show slightly better performance of the stack filter over the Lee filter. This quality assessment is not conclusive but indicates the potential of stack filters in SAR image processing for visual analysis.
The classification results and the quality perception results suggest that stack filters are promising tools in SAR image processing and analysis.

\bibliographystyle{IEEEbib}
\bibliography{bib-maelenaCiarp2011}

\end{document}